\documentclass[conference,letterpaper]{IEEEtran}

\usepackage[T1]{fontenc}
\usepackage{microtype}
\usepackage{amsmath,amssymb}
\usepackage{graphicx}
\usepackage{booktabs}
\usepackage{placeins}
\usepackage{algorithm}
\usepackage{algpseudocode}
\usepackage{cite}
\usepackage[hidelinks]{hyperref}
\title{HHW-Ego: DSLR-Quality Enhancement for Multi-Source Wearable Ego Imaging}
\author{\IEEEauthorblockN{Yunfei Tong, Qianqian SUN, Jixiang Luo, Dell Zhang, Hao Sun}}

\begin{document}
\maketitle

\begin{abstract}
  Wearable ego-centric images are now in high demand for scenarios ranging from daily smart glass usage to embodied intelligence. But the image quality is far behind smart phone due to the lacking paired high-quality reference images and the dedicated enhancement systems. To solve this, a customized degradation pipeline is designed to generate paired samples matching wearable camera images, boosting the upper limit of enhancement performance. Besides a two-stage enhancement framework is further built: first, tuning an efficient model on the paired dataset to enhance real wearable images; second, using hyperspectral data to refine color temperature for better quality. Moreover we present HHW-Ego, a multi-source paired dataset integrating with Hyperspectral, High-dynamic range and Wearable Ego-centric data, which includes 5,000 image groups spanning indoor/outdoor and day/night scenes. On held-out paired data, Stage-1 fine-tuning improves PSNR, SSIM, Lab distance, and LPIPS for Real-ESRGAN and PFT-SR. On real wearable images, no-reference metrics, human preference, gamut analysis, and visual comparisons characterize the effects of Stage~2.
\end{abstract}


\section{Introduction}
\label{sec:intro}

Wearable ego-centric imaging has recently drawn increasing attention due to its wide range of applications, including daily smart-glass usage, life-logging, and embodied intelligence for robotics and AR/VR systems \cite{ Betancourt2015, ragusa2020ego, Ego4D2022, Yang_2025_CVPR}. Compared with traditional third-person cameras, wearable devices offer a first-person perspective that captures unique information about user activities and environments. Such capabilities make them valuable for human–computer interaction, navigation, and downstream embodied intelligence tasks \cite{ZhangPerceptualMetric2018, Broxton2020}. Despite this potential, the visual quality of wearable ego-centric images remains significantly behind that of smartphone or DSLR cameras, which greatly limits their utility in both consumer and research settings.
The primary challenge stems from the intrinsic hardware limitations of wearable cameras. Compact form-factors, limited optics, low dynamic range, and inaccurate color rendition penalize wearable systems \cite{chen2018learning}. Furthermore, unlike smartphones, which benefit from large-scale paired datasets and dedicated imaging pipelines, ego-centric cameras lack high-quality reference images for supervised training. This absence of paired supervision hinders the development of effective enhancement models, leaving most wearable imagery unsuitable for high-level vision tasks or immersive user experiences.

In recent years, there have been significant advances in super-resolution (SR) and image enhancement that are relevant to this problem. For example, SwinIBSR \cite{shi2024swinibsr} proposes a Transformer-based method for real-world infrared image super-resolution, with a novel degradation model and mixed training to better generalize across visible and infrared domains. Physics-enhanced networks have been demonstrated to reconstruct high-resolution images from compressive measurements over expansive scenes \cite{jin2024parallel}. Moreover, in image enhancement under novel optics, researchers have used deep learning to boost image quality of metalens cameras by learning paired low- vs high-quality images to recover fine detail and reduce optical blur \cite{seo2024deep}. These methods show that with the right degradation modeling and paired supervision, SR/enhancement performance can be dramatically improved. HVI~\cite{yan2025hvi} introduces a new color space for low-light improvement, GaussHDR~\cite{liu2025gausshdr} combines 3D Gaussian splatting with HDR reconstruction, and UHD-processer~\cite{liu2025uhd} unifies UHD image restoration. Besides \cite{feijoo2025darkir} focus on the restoration of low-light infrared images. 

In the hyperspectral domain, there have also been promising developments. Deep learning-based hyperspectral image reconstruction (from RGB to HSI) has been used for agri-food quality assessment, demonstrating that reconstructed spectra can approximate ground truth well, enabling downstream tasks \cite{ahmed2024deep}. There is also a comprehensive review surveying hyperspectral image reconstruction methods from compressive measurements, which underscores the importance of recovering both spectral and spatial fidelity \cite{han2025recent}. Additionally, algorithms for fusing high spatial resolution multispectral images with low spatial resolution hyperspectral images (spatial-spectral feature fusion) have been developed to yield high-spatial resolution hyperspectral output \cite{yi2024spatial}. Such methods are relevant because improving color fidelity, spectral consistency, and detail are all critical for wearable ego-centric imaging, especially under challenging illumination.

To address these issues specifically for wearable ego-centric images, we propose a two-stage enhancement framework. First, we design a customized degradation pipeline that synthesizes realistic wearable-like data from high-quality sources, thereby generating paired samples that bridge the gap between reference-rich domains and wearable devices. This enables efficient training of enhancement models on data that faithfully mimics real-world degradations. Second, we leverage hyperspectral data to refine color temperature and achieve perceptually consistent color reproduction, which is particularly critical for scenes with complex illumination. Together, these strategies boost the upper limit of enhancement quality and bring wearable imagery closer to smartphone- and DSLR-level fidelity.
Beyond the framework itself, we introduce HHW-Ego, a novel multi-source paired dataset that integrates Hyperspectral, High-dynamic range (HDR), and Wearable Ego-centric data. The dataset contains 5,000 image groups spanning diverse indoor/outdoor and day/night scenes, offering comprehensive coverage for both supervised enhancement and generalization analysis. We evaluate restoration fidelity on held-out LQ--HDR pairs and assess real wearable images with no-reference metrics, human preference, and gamut analysis. We believe that HHW-Ego, together with our proposed pipeline, will serve as a strong foundation for future research on wearable image enhancement, perception, and embodied intelligence.

Our contributions can be summarized as follows:
\begin{itemize}

\item We identify the fundamental limitations of wearable ego-centric imaging and design a customized degradation pipeline to generate realistic paired training samples.

\item We develop a two-stage enhancement framework that couples efficient supervised training with hyperspectral-guided color refinement.

\item We construct HHW-Ego, the first multi-source paired dataset integrating hyperspectral, HDR, and wearable ego-centric imagery across diverse environments.

\item We evaluate the two stages separately, measuring fidelity on paired data and perceptual quality on real wearable images.
\end{itemize}

\section{Related Work}

CNN-based and GAN-based super-resolution works—represented by BSRGAN~\cite{zhang2021designing} and Real-ESRGAN~\cite{wang2021real}—typically adopt a two-component architecture: CNNs serve as the backbone to extract and reconstruct image features, while GANs (as proposed in~\cite{goodfellow2020generative}) are integrated mainly as a post-processing module or loss function to enhance visual quality. The CNN backbone, with its efficient hierarchical feature extraction capability, lays the foundation for recovering fine details from low-quality images; GANs, meanwhile, optimize the output by minimizing the distribution gap between generated and real high-quality images—often via adversarial loss—to refine texture and contrast.
This design offers notable advantages for industrial scenarios: the CNN-based framework ensures computational efficiency (reducing inference latency), and the GAN-driven optimization avoids over-smoothing common in traditional SR methods. These traits make such models widely applicable in streaming media, real-time video enhancement, and other latency-sensitive tasks where rapid processing is prioritized. However, the introduction of GANs also brings inherent limitations: the adversarial training process may generate unrealistic textures or artifacts (e.g., unnatural edge halos, inconsistent color patches), which, while improving objective metrics like PSNR in some cases, restrict subjective quality—especially for wearable ego-centric images where structural consistency and perceptual naturalness (as evaluated by LPIPS~\cite{zhang2018perceptual}) are critical for downstream tasks like embodied interaction or AR overlay.

However, the practical large-scale application of Transformer- and Diffusion-based super-resolution methods in industry is hindered by prohibitive computational costs, which stem from two core limitations. Firstly, Transformer-induced computational overhead as PASD~\cite{yang2024pixel} and PFT-SR~\cite{long2025progressive}: the self-attention mechanism in Transformer-based modules—adopted in methods like PASD exhibits quadratic time complexity ($O(n^2)$, where n denotes the number of image tokens). This characteristic leads to enormous memory consumption and computational demands when processing high-resolution wearable ego-centric images (e.g., 1080p and above), rendering real-time inference infeasible on resource-constrained wearable devices such as smart glasses equipped with low-power GPUs. Secondly, Diffusion sampling latency like StableSR~\cite{wang2024exploiting}  and SUPIR~\cite{yu2024scaling}: Diffusion-based components, as seen in StableSR and PFT-SR, typically require dozens of iterative sampling steps to generate high-quality enhanced images. Even with step-reduction optimizations, the resulting inference latency remains substantially higher than that of CNN-GAN methods, failing to meet the real-time performance requirements of critical industrial scenarios—including live streaming, real-time AR overlay for smart glasses, and low-latency feedback loops in wearable device-driven embodied intelligence tasks.


\section{Methodology}
\begin{figure*}[tp]
    \centering
    \includegraphics[width=.85\linewidth,height=.54\textheight,keepaspectratio]{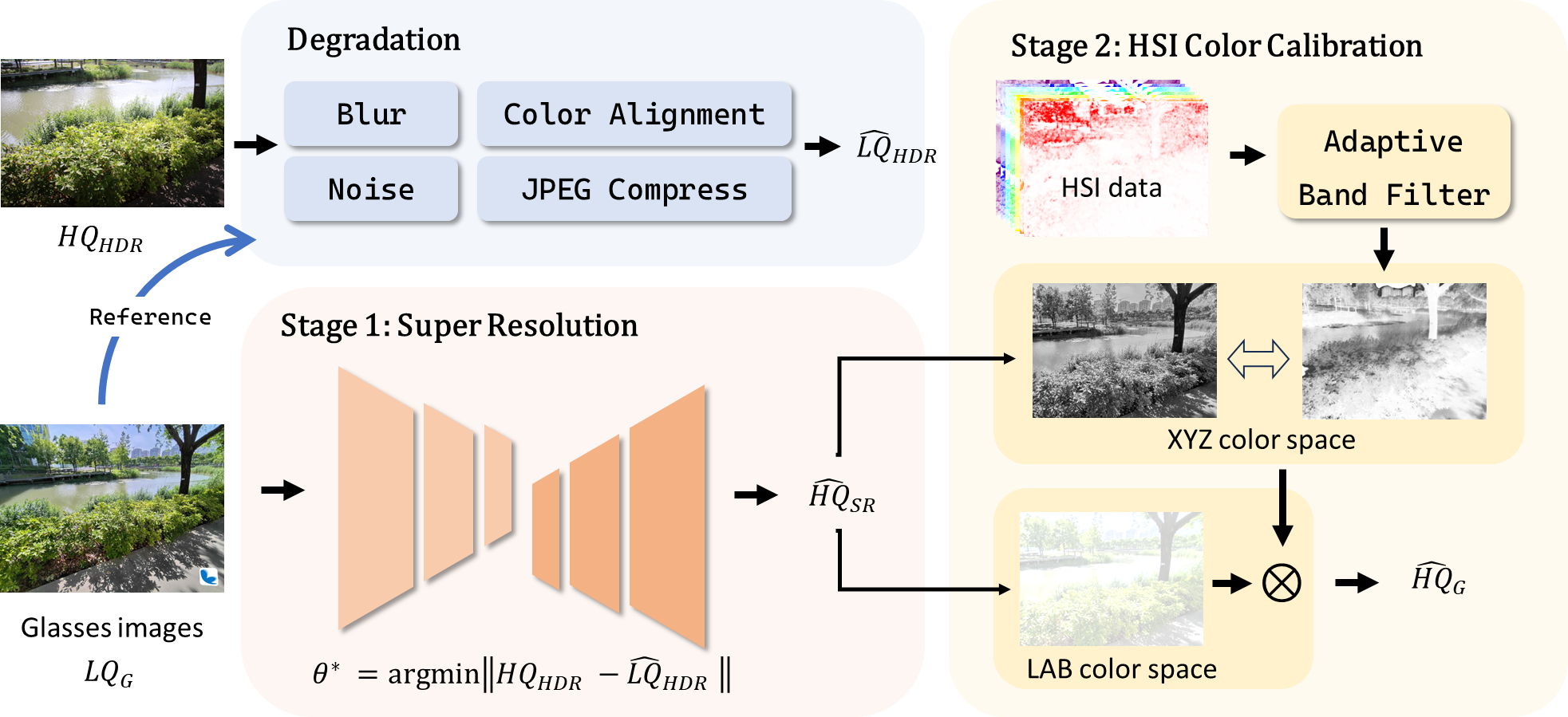}
    \caption{Overview of our two-stage enhancement pipeline. Stage 1 utilizes an aligned degradation model to train a super-resolution network for detail recovery. Stage 2 integrates hyperspectral (HSI) information via XYZ/LAB color space transformation to achieve precise color calibration and overexposure recovery.}
    \label{fig:overreview}
\end{figure*}

\subsection{Overview}

Our data pipeline consists of acquisition, preprocessing, degradation, and enhancement as shown in \textbf{Figure~\ref{fig:cameras} (b)} and \textbf{Figure ~\ref{fig:overreview}}. We captured paired wearable-device images, HDR images, and hyperspectral (HSI) images across over 5000 indoor and outdoor scenes under both daytime and nighttime conditions. Due to differences in camera parameters and viewing angles, we first perform cross-modal image registration to ensure approximate semantic alignment across the three sets.

To enhance wearable-device images, we train a super-resolution reconstruction model capable of enhancing details in low-resolution wearable-style images. To obtain paired low-quality (LQ) and high-quality (HQ) images for training, we degrade HDR images while aligning their style and noise characteristics with wearable-device images in both Lab and RGB color spaces. The resulting paired data is used to train an SR model $\theta $, forming the first stage of enhancement, which not only restores fine details but also adapts the wearable images towards HDR-style quality. In the second stage, we leverage HSI information to further enhance color and detail in the visible spectrum.

\subsection{Aligned Degradation}

Conventional super-resolution models aim to restore high-frequency details that are lost during image downsampling. 
In practice, the degradation process is typically simulated by applying blur and noise to high-quality images, 
where randomized blur kernels and noise levels are often introduced to enhance model robustness as ~\cite{damera2000image, pei2019effects}. 
However, such synthetic degradations rarely reflect the true distribution of real-world wearable images, leading to a mismatch between training and deployment conditions.  

Formally, let the real-world LQ image be denoted as $LQ_{\text{source}}$ 
and the corresponding HQ target as $HQ_{\text{source}}$. 
We aim to learn a mapping $\theta$ such that
\begin{equation}
\theta(LQ_{\text{source}}) \approx HQ_{\text{source}}.
\label{eq:sr_mapping}
\end{equation}
In standard blind SR settings, a degraded image is generated as
\begin{equation}
LQ_{\text{target}} = (HQ_{\text{target}} \otimes k) \downarrow_s + n,
\label{eq:classic_degradation}
\end{equation}
where $k$ denotes the blur kernel, $\otimes$ represents convolution, $\downarrow_s$ indicates downsampling by factor $s$, 
and $n$ is additive noise. This synthetic degradation does not guarantee alignment between $LQ_{\text{target}}$ and the real $LQ_{\text{source}}$.  

To better adapt $\theta$ to wearable imagery, we define an \emph{aligned degradation process}:
\begin{equation}
LQ_{\text{target}} = (\text{Lab}_s(HQ_{\text{target}}) \otimes \hat{k}) \downarrow_s + \hat{n},
\label{eq:aligned_degradation}
\end{equation}
where $\text{Lab}_s(\cdot)$ performs Lab-space style alignment, $\hat{k}$ is a blur kernel estimated to match 
$LQ_{\text{target}}$ with $LQ_{\text{source}}$, and $\hat{n}$ is noise modeled from the source domain.  

Overall, the Blind SR task can be formulated as a maximum a posteriori (MAP) estimation:
\begin{equation}
\arg\max_x \, p(y \mid x, s, k, n)\, p(x)\, p(s, k, n),
\label{eq:map_formulation}
\end{equation}
where $p(x)$ denotes the prior over HR images, and $p(k)$ serves as the alignment prior. 
The aligned degradation model assumes independence among $s$, $k$, and $n$.

\subsection{Two-Stage Multimodal Enhancement}
Enhancement of wearable-device images combines detail reconstruction and color correction in a unified framework. In the first stage, we fine-tune a super-resolution model on paired $HQ_{\text{target}}$ and $LQ_{\text{target}}$ images obtained from the aligned degradation, enabling the model to restore fine structures while adapting to the style of the wearable devices. In the second stage, we leverage hyperspectral (HSI) data to further refine colors in the visible range by selecting informative spectral bands and applying correlated color temperature (CCT) mapping, producing enhanced images that are both detailed and color-consistent with the source scenes.

\subsubsection{Hyperspectral Band Selection}












Hyperspectral and RGB images often suffer from illumination color casts, spectral distortions, and overexposed regions. To enhance color fidelity, we perform a three-stage correction pipeline: white balance correction, HSI color correction, and overexposure recovery. Given an image cube $I \in \mathbb{R}^{H \times W \times C}$, each pixel spectrum $s_i \in \mathbb{R}^C$ is processed individually or in local neighborhoods.

\textbf{White Balance Correction.} Given an HSI or RGB image $I$, we first estimate the scene illuminant $\mathbf{L} = [L_R, L_G, L_B]$ by averaging high-intensity pixels or using a reference white region. The white-balanced image is computed as
\begin{equation}
I'_c = \frac{I_c}{L_c} \cdot \bar{L},
\quad c \in {R,G,B},
\label{eq:white_balance}
\end{equation}
where $\bar{L}$ is the channel-wise normalization of $L_c$. This reduces color casts caused by illumination differences.

\textbf{HSI Color Correction.} For each pixel spectrum $s_i \in \mathbb{R}^C$, a target color reference $\hat{s}i$ is defined based on statistical matching or histogram alignment. The corrected spectrum is obtained by
\begin{equation}
s'i = (s_i - \mu_s) \cdot \frac{\sigma{\hat{s}}}{\sigma_s} + \mu{\hat{s}},
\label{eq:hsi_color_correction}
\end{equation}
where $(\mu_s, \sigma_s)$ and $(\mu_{\hat{s}}, \sigma_{\hat{s}})$ denote the mean and standard deviation of the source and reference bands, respectively. This operation compensates for color drifts across the spectral channels.

{\textbf{Overexposure Recovery.} For each pixel, an overexposure mask $M$ is defined as
\begin{equation}
M =  \text{max}(I_X, I_Y, I_Z) > 1.0,
\label{eq:overexposure_mask}
\end{equation}
where $I_X, I_Y, I_Z$ are the XYZ channels. The mask $M$ is then dilated to form an outer ring $R$, and only sufficiently bright pixels are retained:
\begin{equation}
R_{\text{corrected}} = \text{mean}(I*R).
\label{eq:overexposure_ring}
\end{equation}
Finally, pixel values in $R_{\text{corrected}}$ are adjusted using local neighborhood statistics or interpolation from valid neighboring pixels to restore saturated colors.}

\subsubsection{HSI to CCT-Guided RGB Correction}

The proposed algorithm performs an efficient CCT-based color correction by jointly leveraging hyperspectral and RGB information. First, a subset of hyperspectral bands is converted into the CIE XYZ color space, from which a pixel-wise CCT map is estimated. This CCT map serves as the target illumination reference, effectively capturing the fine-grained spectral characteristics of the hyperspectral data. In parallel, the input RGB image is converted to XYZ and its original CCT is computed. By taking the ratio between the target and original CCTs, a correction factor is derived and adaptively applied to the XYZ representation of the RGB image. This adjustment efficiently transfers illumination consistency from the hyperspectral domain to the RGB domain, achieving robust color fidelity. Finally, the corrected XYZ is transformed back into RGB space. Through this procedure, the algorithm provides an effective fusion strategy that exploits hyperspectral priors to enhance RGB imagery with minimal computational overhead. And the corresponding algorithm is as following to get $I_{\text{rgb\_corrected}}$: 

\begin{algorithm}[H]
\caption{CCT-based Color Correction from HSI to RGB}
\label{alg:cct_correction}
\begin{algorithmic}[1]
\Require HSI bands $H_{\text{sub}}$,  RGB image $I_{\text{rgb}}$, correction strength $\text{strength}$
\Ensure Corrected RGB image $I_{\text{rgb\_corrected}}$
\State Convert $H_{\text{sub}}$ to CIE XYZ using standard color matching functions in ~\cite{ccup1931commission}
\State Compute pixel-wise CCT map of $T_{\text{target}}$
\State Convert $I_{\text{rgb}}$ to XYZ and compute original CCT of $T_{\text{original}}$
\State Compute CCT adjustment ratio for each pixel $p$:
\[
R_{\text{cct}}(p) = \left(\frac{T_{\text{target}}(p)}{T_{\text{original}}(p)}\right)^{\text{strength}}
\]
\State Adjust XYZ channels by $R_{\text{cct}}$:
\[
XYZ_{\text{adj}} = XYZ_{\text{rgb}} \cdot R_{\text{cct}}
\]
\State Convert $XYZ_{\text{adj}}$ back to RGB:
\[
I_{\text{rgb\_corrected}} = \text{XYZ\_to\_RGB}(XYZ_{\text{adj}})
\]
\State Clip $I_{\text{rgb\_corrected}}$ into $[0,1]$ and save result
\end{algorithmic}
\end{algorithm}

\section{Dataset}
\subsection{Multi-Source Data Collection}
\begin{figure*}[tp]
    \centering
    \includegraphics[width=0.9\linewidth,height=.54\textheight,keepaspectratio]{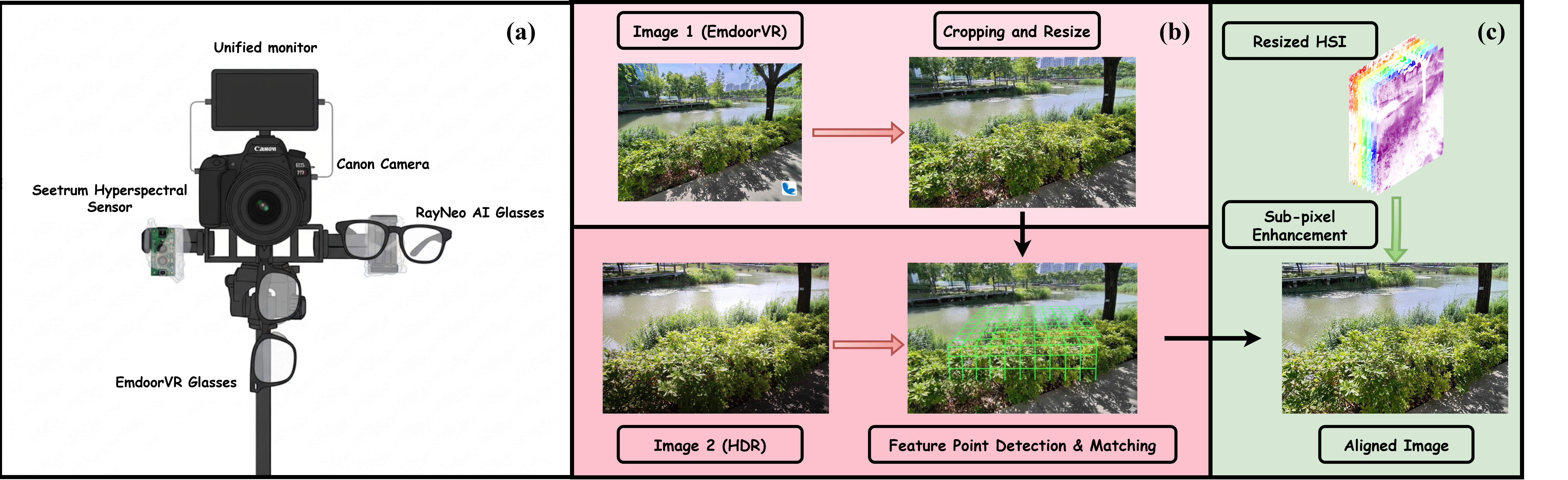}
    \caption{(a) Schematic of the HHW-Ego data acquisition platform. (b) The image co-registration workflow, including cropping, resizing, and feature-based matching. (c) The sub-pixel target enhancement pipeline integrating resized HSI information for the final output.}
    \label{fig:cameras}
\end{figure*}

\begin{table*}[tp]
\caption{Multi-source camera characteristics.}
\label{tab:data_collection}
\centering
\renewcommand{\arraystretch}{1.15}
\begin{tabular}{lcc}
\toprule
\textbf{Devices} & \textbf{Raw Size} & \textbf{Data Description} \\
\midrule 
EmdoorVR Glasses             & $4000 \times 3000\times3$  &  RGB(8bit) \\
RayNeo V3 AI Glasses      &  $3864_{\pm168} \times 2760_{\pm264} \times3$  &  RGB(8bit)  \\
Canon EOS 77D DSLR         & $6000\times4000\times3$  &   HDR(14bit)\\
Seetrum Hyperspectral Sensor  & $ 1184\times 1600 \times31$  & 380 - 980 nm   \\
\bottomrule
\end{tabular}
\end{table*}

To address the inherent limitations of imaging quality in Wearable selfie devices (such as smart glasses), we constructed the HHW-Ego (Hyperspectral-HDR-Wearable Ego-centric dataset), which is a multimodal and multi-source paired dataset for wearable visual enhancement tasks with varying data description as shown in Table~\ref{tab:data_collection}. HHW-ego integrates HSI, HDR, and wearable first-person perspective (ego-centric) images, aiming to provide high-quality reference signals and support end-to-end enhancement training from low-quality wearable images to high-quality perception-consistent images . 

\textbf{Wearable Device Images:} The Qualcomm AR1 platform paired with SONY IMX681 sensor is a mainstream high-end smart glasses solution. It processes images through a complex image signal  pipeline(ISP), converting Bayer's raw data into the final polished RGB image. We use current consumer-grade wearable vision systems EmdoorVR Glasses \footnote{https://vr.emdoor.com/} and RayNeo AI Glasses \footnote{https://www.rayneo.cn} in \textbf{Figure \ref{fig:cameras}} to capture the corresponding images.

\textbf{High Dynamic Range Image (HDR):} Use professional-grade single-lens reflex cameras (such as Canon EOS 77D) to capture multiple frames of exposure images at the same time point and from the same perspective, and generate high dynamic range images through HDR synthesis algorithms. HDR images retain rich details from dark to bright areas, serving as high-quality visible light domain references for subsequent super-resolution and detail reconstruction tasks.

\textbf{Hyperspectral Image (HSI):} The spectral cube data of the scene was collected using a snapshot spectral imager (Seetrum Hyperspectral Sensor) \footnote{https://www.seetrum.com/}. Each pixel contains a continuous spectral response within the 380-980 nm band range, providing rich physical color information.
All acquisition processes adopt a fixed tripod + synchronous trigger mechanism to ensure the spatial consistency of the three modal images. Meanwhile, detailed metadata such as timestamps, light intensity, camera parameters (focal length, shutter speed), ambient temperature and geographical location should be recorded to facilitate subsequent analysis and modeling.
\subsection{Image Registration}
To bridge the spatial and spectral gaps between heterogeneous sensors, we developed a specialized processing pipeline as illustrated in Figure \ref{fig:cameras}. Due to inherent disparities in field of view (FOV), focal length, and geometric distortion among the sensors, a multi-stage fine cross-modal registration is performed post-acquisition.

As shown in Figure \ref{fig:cameras}(b), the low-quality (LQ) image from the EmDoorVR wearable device is first cropped and resized to match the reference HDR image. This stage defines a common overlapping region and standardizes the resolution for subsequent fusion. We then employ the Scale-Invariant Feature Transform (SIFT) to detect robust keypoints and establish correspondences between the LQ and HDR modalities. By estimating an optimal transformation matrix (e.g., affine or perspective projection), the HDR and hyperspectral (HSI) data are precisely warped into the LQ coordinate system, achieving pixel-level spatial alignment.

Simultaneously, as depicted in Figure \ref{fig:cameras}(c), the hyperspectral cubes are spatially resampled and integrated to provide dense spectral cues. We perform sub-pixel enhancement by fusing the resized HSI information with the aligned visual frames, ensuring high-fidelity color and texture reconstruction. The final HHW-Ego dataset comprises 5,000 sets of strictly aligned image triplets (LQ, HQ, and HSI), and captures diverse indoor and outdoor environments across varying lighting conditions (e.g., day and night), ensuring high generalization for real-world wearable applications. These sequences maintain rigorous consistency across spatial, semantic, and pixel-level dimensions, providing a robust foundation for multimodal fusion, super-resolution, and spectral enhancement tasks in embodied intelligence scenarios.
\section{Experiments}
\subsection{Experimental Setup and Evaluation Protocol}
We evaluate Stage~1 on held-out LQ--HDR pairs, Stage~2 on real wearable images, and device-aligned degradation by measuring the distance between synthesized LQ and real wearable distributions. Each Stage-1 input is synthesized from its HDR reference, enabling full-reference evaluation with PSNR, SSIM~\cite{wang2004image}, Lab distance, and LPIPS~\cite{zhang2018perceptual}. Real wearable, HDR, and HSI captures use different optical systems and lack pixel-aligned ground truth for full-reference evaluation. We therefore evaluate Stage~2 with seven no-reference metrics (NIQE~\cite{mittal2012making}, BRISQUE~\cite{mittal2012no}, PIQE~\cite{venkatanath2015blind}, MUSIQ, CLIPIQA, NIMA, and DBCNN), visual comparisons, human preference, and gamut analysis.

We estimate blur kernels with KernelGAN~\cite{KernelGAN} and add Gaussian--Poisson noise. Gaussian parameters are estimated from wearable images and Poisson parameters are sampled from $[0.02, 0.1]$. The Real-ESRGAN generator is fine-tuned with an $\mathcal{L}_1$ objective and Adam for 1,000 iterations at a learning rate of 0.1. Fine-tuning and inference use an NVIDIA A100 GPU. The HSI-guided correction is training-free. Compared models use official pretrained weights unless marked \emph{+ FT}. In the real-image comparison, \emph{+ Ours} applies Stage-2 correction to each baseline output without retraining the baseline.

\subsection{Stage-1 Paired Restoration Evaluation}
\begin{table}[!tbp]
\caption{Stage-1 full-reference performance on held-out LQ--HDR pairs. \emph{+ FT} denotes fine-tuning on device-aligned paired data.}
\label{tab:stage1_paired}
\centering
\small
\setlength{\tabcolsep}{2.5pt}
\begin{tabular}{@{}lcccc@{}}
\toprule
Model & PSNR $\uparrow$ & SSIM $\uparrow$ & Lab distance $\downarrow$ & LPIPS $\downarrow$ \\
\midrule
Real-ESRGAN & 25.7492 & 0.7753 & 0.2937 & 0.3294 \\
+ FT & \textbf{30.0134} & \textbf{0.8589} & \textbf{0.1384} & \textbf{0.1925} \\
PFT-SR & 32.2266 & 0.8678 & 0.3004 & 0.4762 \\
+ FT & \textbf{33.6825} & \textbf{0.8891} & \textbf{0.1514} & \textbf{0.3870} \\
\bottomrule
\end{tabular}
\end{table}
Fine-tuning improves all four metrics for both backbones (Table~\ref{tab:stage1_paired}). PSNR rises by 4.26~dB for Real-ESRGAN and 1.46~dB for PFT-SR. SSIM increases, while Lab distance and LPIPS decrease for both backbones.

\subsection{Stage-2 Evaluation on Real Wearable Images}
\begin{figure*}[tp]
    \centering
    \includegraphics[width=0.95\linewidth,height=.54\textheight,keepaspectratio]{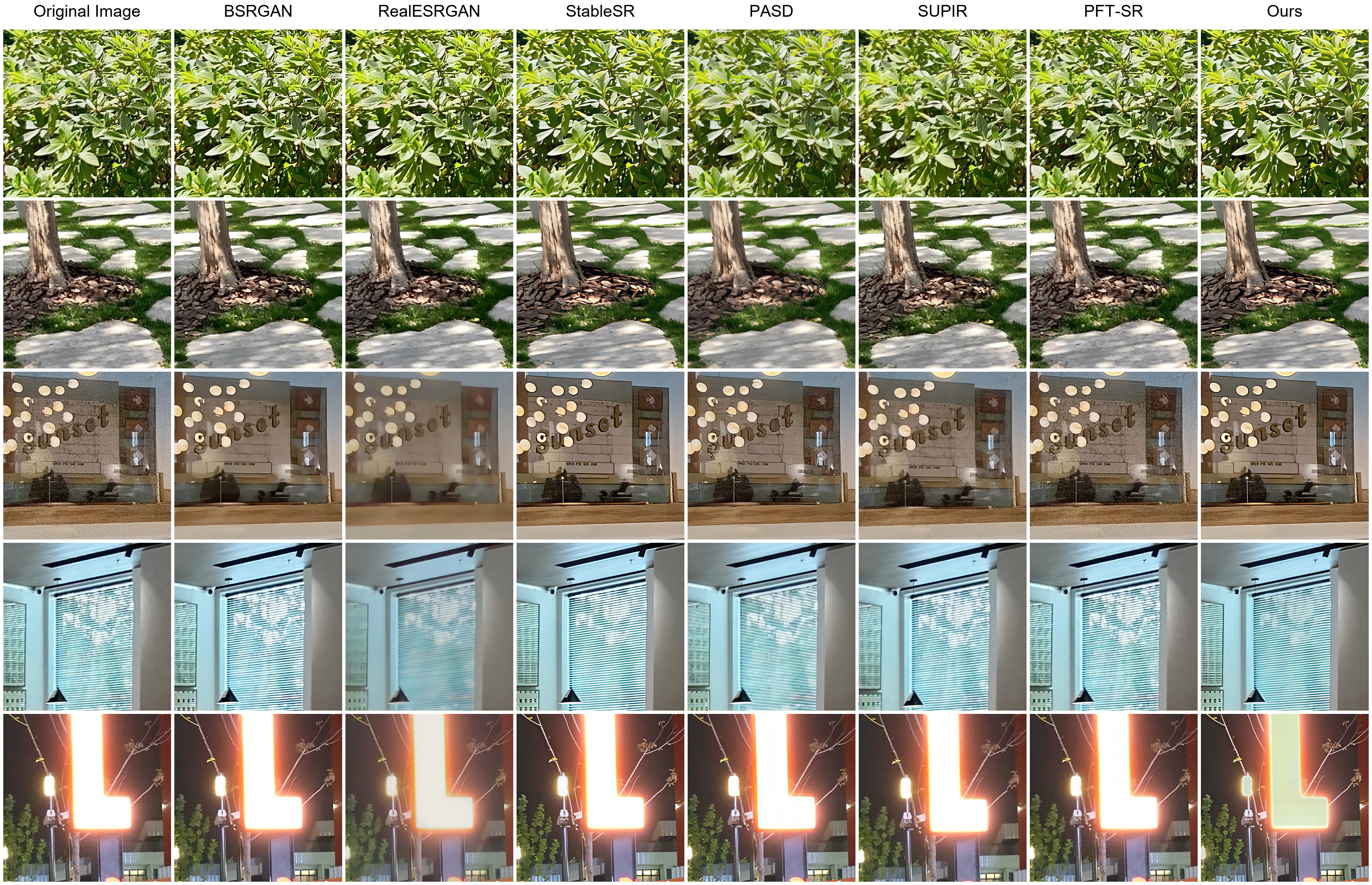}
    \caption{Visual comparison with state-of-the-art super-resolution methods on diverse real-world scenes.}
    \label{fig:comp-SOTA}
\end{figure*}

\begin{figure*}[tp]
    \centering
    \includegraphics[width=0.95\linewidth,height=.54\textheight,keepaspectratio]{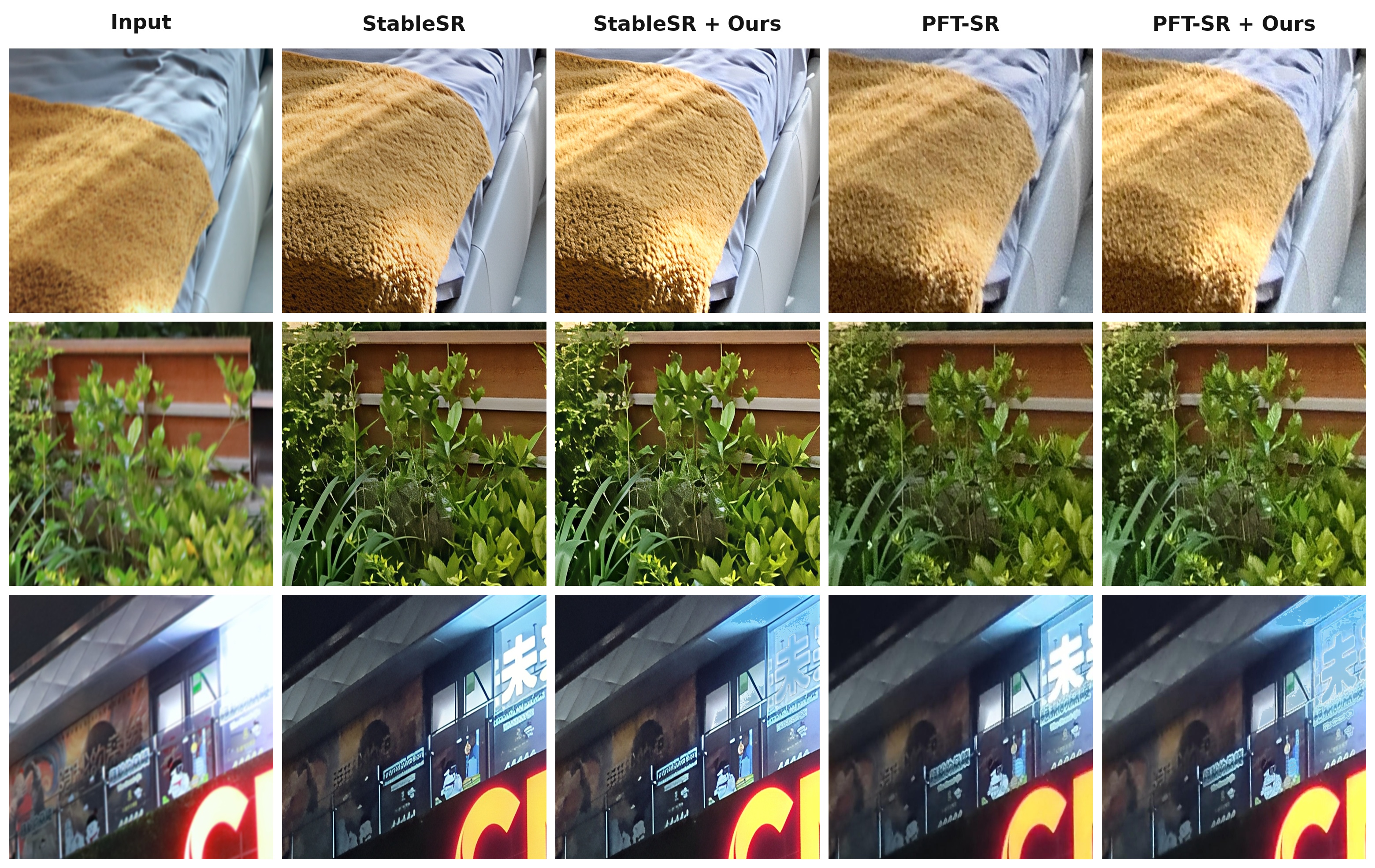}
    \caption{Additional visual comparison of restoration methods on real wearable images.}
    \label{fig:main-demo}
\end{figure*}

\begin{table*}[tp]
\centering
\caption{Effect of Stage-2 HSI-guided correction on six restoration backbones for real wearable images}
\label{tab:comp_metrics}
\resizebox{.90\textwidth}{!}{
\begin{tabular}{lccccccc}
\toprule
Method & NIQE $\downarrow$ & BRISQUE $\downarrow$ & PIQE $\downarrow$ & MUSIQ $\uparrow$ & CLIPIQA $\uparrow$ & NIMA $\uparrow$ & DBCNN $\uparrow$ \\
\midrule
BSRGAN~\cite{zhang2021designing} & 5.3877 & 36.2413 & 42.7953 & 55.9860 & 0.5396 & 4.4188 & 0.4906 \\
+ Ours & \textbf{4.9118} & \textbf{30.7168} & \textbf{39.7603} & \textbf{58.8088} & \textbf{0.5814} & \textbf{4.5815} & \textbf{0.5520} \\
\midrule
RealESRGAN~\cite{wang2021real} & 5.2948 & 30.3596 & 42.0978 & 55.2945 & 0.4812 & 4.3306 & 0.5038 \\
+ Ours & \textbf{5.0188} & \textbf{27.5894} & \textbf{39.6600} & \textbf{55.9863} & 0.4781 & \textbf{4.4105} & \textbf{0.5265} \\
\midrule
PASD~\cite{yang2024pixel} & 5.3925 & 20.8090 & 32.4830 & 64.9594 & 0.4586 & 4.5107 & 0.6124 \\
+ Ours & \textbf{5.2152} & \textbf{20.2964} & 32.9309 & 64.7741 & \textbf{0.4656} & \textbf{4.5196} & \textbf{0.6189} \\
\midrule
StableSR~\cite{wang2024exploiting} & 5.8013 & 24.8340 & 26.6623 & 61.3339 & 0.4456 & 4.3201 & 0.5502 \\
+ Ours & \textbf{5.6274} & 27.6523 & 28.2251 & \textbf{62.1308} & \textbf{0.4831} & \textbf{4.3788} & \textbf{0.5719} \\
\midrule
SUPIR~\cite{yu2024scaling} & 4.5177 & 13.4245 & 24.5585 & 51.5649 & 0.4307 & 4.3054 & 0.5036 \\
+ Ours & \textbf{4.3152} & 14.9134 & 27.0041 & \textbf{54.6477} & \textbf{0.4840} & \textbf{4.4104} & \textbf{0.5523} \\
\midrule
PFT-SR~\cite{long2025progressive} & 6.6805 & 49.5250 & 61.0048 & 37.1002 & 0.3359 & 3.6841 & 0.3201 \\
+ Ours & \textbf{5.8655} & \textbf{46.4815} & \textbf{58.7615} & \textbf{42.8771} & \textbf{0.3425} & \textbf{3.8779} & \textbf{0.4236} \\
\bottomrule
\end{tabular}
}
\end{table*}
The HSI-guided correction improves 35 of 42 backbone--metric pairs (Table~\ref{tab:comp_metrics}). NIQE, NIMA, and DBCNN improve for all six backbones, while MUSIQ and CLIPIQA improve for five. BSRGAN improves on all seven metrics, and PFT-SR gains 5.77 points in MUSIQ. Figs.~\ref{fig:comp-SOTA} and~\ref{fig:main-demo} show qualitative comparisons. The correction targets color casts and overexposure while preserving the structure of the baseline outputs.

\subsection{Ablation Studies}
\begin{table}[!tbp]
\caption{Component-wise degradation ablation.}
\label{tab:degradation_ablation}
\centering
\footnotesize
\setlength{\tabcolsep}{2.5pt}
\begin{tabular}{@{}lllcccc@{}}\hline
\multicolumn{1}{c}{\bf Color} & \multicolumn{1}{c}{\bf Blur} & \multicolumn{1}{c}{\bf Noise} & \multicolumn{1}{c}{\bf LPIPS $\downarrow$} & \multicolumn{1}{c}{\bf FID $\downarrow$} & \multicolumn{1}{c}{\bf SSIM $\uparrow$} & \multicolumn{1}{c}{\bf PSNR $\uparrow$} \\ \hline
None & Random & Random & 0.2239 & 51.8279 & 0.4269 & 13.9433 \\
Aligned & Random & Random & 0.2201 & 51.1615 & 0.4244 & 14.0457 \\
Aligned & Estimated & Random & 0.2202 & 50.9877 & 0.4262 & 14.0682 \\
Aligned & Estimated & Aligned & 0.2202 & \textbf{50.9031} & 0.4258 & 14.0620 \\\hline
\end{tabular}
\end{table}
\textbf{Degradation components.} Color alignment lowers LPIPS and FID relative to random blur and noise with no color alignment. Estimating the blur kernel further lowers FID (Table~\ref{tab:degradation_ablation}). Night lighting, motion, and overexposure introduce different mixtures of artifacts, so the optimal configuration can vary by condition.

\begin{table}[!tbp]
\caption{Stage-2 module ablation on real wearable images. Lower is better for all metrics.}
\label{tab:ablation-quality_metrics}
\centering
\small
\setlength{\tabcolsep}{2.5pt}
\begin{tabular}{@{}lccc@{}}
\toprule
Method & NIQE $\downarrow$ & BRISQUE $\downarrow$ & PIQE $\downarrow$ \\
\midrule
SR & 6.1946 & 24.0594 & \textbf{53.9674} \\
SR + HSI & 6.1307 & 24.0953 & 55.6191 \\
SR + HSI + Exposure & \textbf{6.0919} & \textbf{23.5923} & 55.2639 \\
\bottomrule
\end{tabular}
\end{table}
\textbf{Stage-2 modules.} In Table~\ref{tab:ablation-quality_metrics}, NIQE improves from 6.1946 to 6.0919 across the three configurations. BRISQUE reaches its best value after exposure correction. PIQE rises after adding HSI, then partially recovers after exposure correction and remains above the SR baseline.

\begin{table}[!tbp]
\caption{Distances between synthesized LQ and real wearable distributions under different degradation strategies. Lower is better.}
\label{tab:degradation_distance}
\centering
\footnotesize
\setlength{\tabcolsep}{2.5pt}
\begin{tabular}{@{}p{.27\columnwidth}ccccc@{}}
\toprule
Strategy & Lab $\downarrow$ & Saturation $\downarrow$ & Sharpness $\downarrow$ & Noise $\downarrow$ & FD $\downarrow$ \\
\midrule
HDR original & 3.501 & 6.225 & 126.8 & 0.6022 & 95.69 \\
Random degradation & 3.525 & 6.593 & 309.7 & 1.7470 & 101.9 \\
\shortstack[l]{Statistical noise +\\random color} & 7.771 & 14.25 & 98.30 & 0.6478 & 858.1 \\
\shortstack[l]{Statistical noise +\\statistical color} & \textbf{0.927} & \textbf{2.513} & \textbf{69.18} & \textbf{0.4442} & \textbf{10.95} \\
\bottomrule
\end{tabular}
\end{table}
\begin{figure*}[tp]
    \centering
    \includegraphics[width=0.95\linewidth,height=.54\textheight,keepaspectratio]{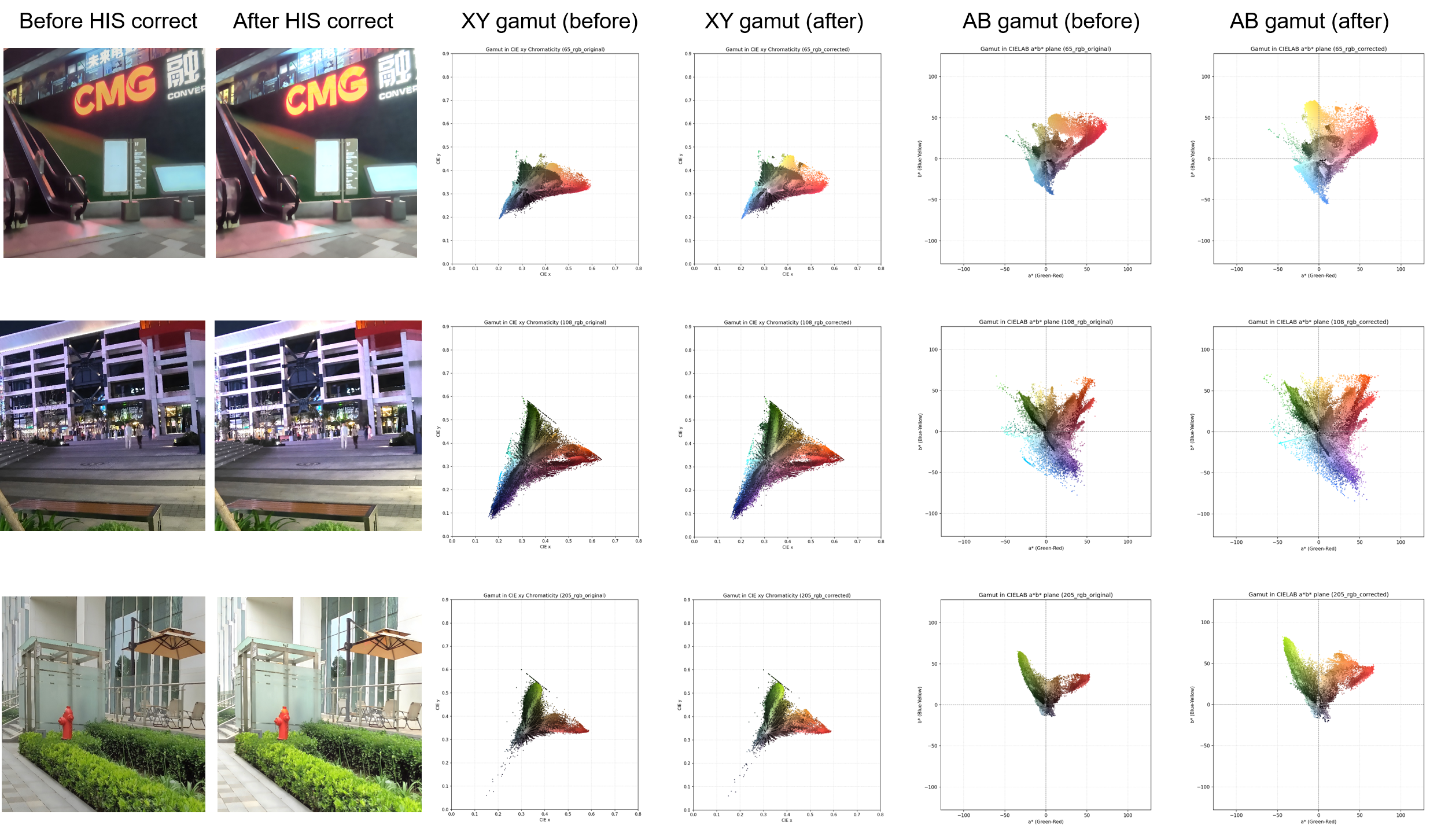}
    \caption{Representative $xy$ and Lab $ab$ gamut distributions before and after HSI-guided correction.}
    \label{fig:comp-gamut}
\end{figure*}

\textbf{Distribution distance.} Table~\ref{tab:degradation_distance} reports distances to real wearable images in Lab statistics, saturation, sharpness, noise, and feature distribution (FD). The device-aligned strategy combines estimated blur, statistical color alignment, noise modeling, and compression. It yields the lowest distance on all five measures and reduces FD from 101.9 with random degradation to 10.95.

\subsection{Human Preference and Gamut Analysis}
We evaluated subjective image quality in a pairwise study with 25 participants. For the same wearable inputs, participants compared the complete method with BSRGAN, Real-ESRGAN, StableSR, PASD, SUPIR, and PFT-SR. The maximum pairwise preference for our method was 93\%.

Fig.~\ref{fig:comp-gamut} shows the $xy$ chromaticity and Lab $ab$ distributions before and after HSI-guided correction. The $xy$ plots show broader chromaticity coverage. Red and blue light sources are more distinct in night scenes, and outdoor vegetation shifts from a yellow-green cluster toward a more natural green distribution. The Lab plots show a less compressed and more balanced distribution.

\FloatBarrier
\section{Conclusion}
We introduce HHW-Ego, a dataset of 5,000 image groups spanning hyperspectral, HDR, and wearable egocentric imagery. Device-aligned degradation generates paired training data for super-resolution, and HSI guides color and exposure correction on real wearable inputs. Fine-tuning improves PSNR, SSIM, Lab distance, and LPIPS for both backbones on held-out Stage-1 pairs. For Stage~2, no-reference metrics, human preference, and qualitative comparisons show improvements on real captures. The module ablation shows that these gains vary by metric. Future work will extend HHW-Ego to more extreme conditions and explore real-time wearable inference.

\bibliographystyle{IEEEtran}
\bibliography{main}

\end{document}